\documentclass[conference]{IEEEtran}

\newtheorem{algorithm}{Algorithm}
\newtheorem{remark}{Remark}

\newtheorem{problem}{Problem}
\newtheorem{proposition}{Proposition}

\usepackage{algorithm}
\usepackage{algorithmic}

\usepackage{csquotes}
\usepackage{dirtytalk}
\usepackage{algorithm,algorithmic}
\usepackage{amssymb}
\usepackage{graphicx}
\usepackage[cmex10]{amsmath}
\usepackage{array}
\usepackage{mdwtab}
\usepackage{fixltx2e}
\usepackage{graphics}
\usepackage{epsfig}
\usepackage{times}
\usepackage{amsmath}
\usepackage{amssymb}
\usepackage{url}
\usepackage{lscape}
\usepackage{color}
\usepackage{epsfig}
\usepackage{cite}
\usepackage{balance}
\usepackage{dblfloatfix}
\usepackage{hyperref}
\usepackage{caption}
\usepackage{subcaption}
\usepackage{times}
\usepackage{multicol}
\usepackage[numbers]{natbib}

\begin{document}

\title{Toward Safe and Efficient Human\textendash Robot Interaction via Behavior-Driven Danger Signaling}

\author{
\IEEEauthorblockN{Mehdi Hosseinzadeh}
\IEEEauthorblockA{Department of Electrical\\ and Systems Engineering\\ Washington University in St. Louis\\ St. Louis, MO 63130, USA\\
Email: mehdi.hosseinzadeh@ieee.org}\\  
\and
%<------ Line breaks in the current column
\IEEEauthorblockN{Bruno Sinopoli}
\IEEEauthorblockA{Department of Electrical\\ and Systems Engineering\\ Washington University in St. Louis\\ St. Louis, MO 63130, USA\\
Email: bsinopoli@wustl.edu}
\and
\IEEEauthorblockN{Aaron F. Bobick}
\IEEEauthorblockA{Department of Computer\\ Science and Engineering\\ Washington University in St. Louis\\ St. Louis, MO 63130, USA\\
Email: afb@wustl.edu}
}

\maketitle

\begin{abstract}
This paper introduces the \textit{notion of danger awareness} in the context of Human-Robot Interaction (HRI), which decodes whether a human is aware of the existence of the robot, and illuminates whether the human is willing to engage in enforcing the safety. This paper also proposes a method to quantify this notion as a single binary variable, so-called \textit{danger awareness coefficient}. By analyzing the effect of this coefficient on the human's actions, an online Bayesian learning method is proposed to update the belief about the value of the coefficient. It is shown that based upon the danger awareness coefficient and the proposed learning method, the robot can build a predictive human model to anticipate the human's future actions. In order to create a communication channel between the human and the robot, to enrich the observations and get informative data about the human, and to improve the efficiency of the robot, the robot is equipped with a \textit{danger signaling} system. A predictive planning scheme, coupled with the predictive human model, is also proposed to provide an efficient and Probabilistically safe plan for the robot. The effectiveness of the proposed scheme is demonstrated through simulation studies on an interaction between a self-driving car and a pedestrian.
\end{abstract}

\IEEEpeerreviewmaketitle

%%%%%%%%%%%%%%%%%%%%%%%%%%%%%
\section{Introduction}\label{sec:introduction}
The aim in Human-Robot Interaction (HRI) is to enable efficient and safe interaction among all participating agents. In general, this is a very challenging task, as safety enforcement techniques depend on some presumptions and assumptions that will not necessarily be true in practice, which may hamper the efficiency of the robot. More precisely, robots will inevitably encounter incomplete and possibly erroneous knowledge of the environment and other agents; humans in particular. For instance, a human might be less attentive than normal when facing a self-driving car, presuming that the self-driving car will undertake the safety satisfaction. Thus, robots must be able to safely and timely reason over the uncertainty of the environment they operate in to maintain their efficiency.

Reasoning in uncertain environments is an area where humans excel. Inspired by this, this paper uses models of human decision-making from cognitive science to develop a framework that enables robots to reason over the uncertainties inherent in predicting the actions of humans to enforce safety.

In particular, this paper introduces the notion of danger awareness in HRI. This notion can be used to decode whether a human is aware of the existence of other agents and possible dangers that exist in the environment, and to explicate whether the human is willing to engage in enforcing the safety. By the term danger we refer to a situation in which the likelihood of a collision between the human and the robot is greater than a threshold value if neither the robot nor the human change their behavior. This notion plays an important role in humans daily interaction. For instance, a driver pays more attention to a biker in front of the car than the one behind the car, as the driver thinks that the biker in front of the car is not aware of the car approaching from behind. Another example is a driver who slows down when driving in an area wherein kids are playing a game, while he/she drives at the same speed when adults are playing the same game, because he/she has learnt that kids may jump on the street unawares. These two examples show that humans act based on their (unconscious) reasoning about each other's danger awareness level in safety-demanding interaction. Therefore, the notion of danger awareness should be incorporated in any approaches which aim at enabling the robots to reach the human-level intelligence, and in any safety enforcing schemes.

This paper proposes a method to quantify the notion of danger awareness. More precisely, it is shown that we can model the effect of this notion on human's decisions via a binary variable, so-called \textit{danger awareness coefficient}. This paper also proposes a method in which the robot continually learns the value of this coefficient based upon real-time observations. A planning scheme is also proposed to provide a probabilistically safe plan for the robot. Note that by a probabilistically safe plan we mean a plan where the probability of collision between the human and the robot is less than a certain value.

Despite the current trend in robotics literature (e.g., \cite{Brooks2019,Shamouty2020,Keil2020,Peddi2020}), we believe that it is implausible to learn the human's behavior by only \textit{passively} observing his/her states and actions. Indeed, the human's trajectory may not encode sufficient information about the human. Thus, any planner developed based upon such learning might be tremendously inaccurate, leading to a conservative solution. One possible way to address this issue, and consequently to improve the efficiency of the robot, is to enable the robot to influence the environment to enrich the observations. To cater this need, we assume that the robot is equipped with a danger signaling system. This system can create a communication channel between the human and the robot, where the robot can convey a message to the human and receives the human's reply manifested in his/her actions. More precisely, by means of the danger signal, the robot can \textit{actively} aptly perturb the environment so that the bird's-eye view of the human's behavior observed by the robot is rich enough to reason about the human's opinion on the cooperative safety enforcement. It is noteworthy that the idea of \textit{perturb\&observe} is taken from human interaction as well. A demonstration is a driver who sounds the horn to alert a pedestrian to the danger, and then reasons about his/her consideration by observing his/her actions.

Two key contributions of this paper are: 1) to introduce and quantify the notion of danger awareness, to investigate its effect in HRI safety, and to propose a real-time method to learn its value, and 2) to propose a state-of-the-art planning scheme to provide probabilistically safe plans by taking into account the notion of danger awareness. The main features of this work are: 1) it is general and can be applied to any HRI in which the human can assess the danger and can cooperate with the robot to enforce safety; 2) the robot communicates with the human, which can improve learning performance and efficiency; and 3) the proposed scheme is modular, meaning that any other objective functions (possibly representing other behavioral attributes of the human) can be incorporated into the scheme without changing the structure.

The remainder of this paper is organized as follows. Section \ref{sec:RW} discusses selected related work. Section \ref{sec:ProblemStatement} formulates the problem. Section \ref{sec:PredictiveHumanModel} discusses how to build a predictive human model, how to learn from human's actions, and how to predict human's state in the future. In Section \ref{sec:framework} a scheme for a safe and efficient planning is proposed. Section \ref{sec:simulation} verifies the proposed method through intensive simulation studies. Finally, Section \ref{sec:conclusion} concludes the paper and discusses future work.

\paragraph*{Notation}
We denote the set of real numbers by $\mathbb{R}$, the set of positive real numbers by $\mathbb{R}_{>0}$, and the set of non-negative real numbers by $\mathbb{R}_{\geq0}$. We use $\mathcal{N}(\mu,\sigma)$ to indicate the Gaussian distribution with mean value $\mu$ and covariance $\Sigma$. We denote proportionality by $\propto$, and the transpose of matrix $A$ by $A^\top$.

%%%%%%%%%%%%%%%%%%%%%%%%%%%%%
\section{Related Work}\label{sec:RW}
In recent years, there have been several studies on predicting the actions of humans in the context of HRI. In some work (e.g., \cite{Huber2010}), it is assumed that the robot has complete knowledge about the environment. However, this assumption may not be reasonable in real-world scenarios due to uncertainties in the human's behavior.

As a result, many researchers have focused on developing a method to enable the robots to use the history of the humans' actions and states to predict future actions. In \cite{Shi2004}, propagation networks have been utilized to detect partially ordered sequential actions of the humans. In \cite{Albanese2008}, the authors introduced the concept of constrained probabilistic Petri nets and showed how this concept can be used to predict the actions of humans. In \cite{Kinugawa2017}, Gaussian mixture distribution techniques have been used to model the actions of humans and to predict the timing. Markov models \cite{Vasquez2009,Morris2011,Ding2011} have been used in a variety of studies to predict the timing of the actions of humans. In \cite{Amor2014}, an interaction primitive framework for predicting the most likely future movements of a human is developed. The anticipatory temporal conditional random fields have been used in \cite{Koppula2013} to predict the future actions of the human. Some \textit{ad hoc} methods have also been proposed in the literature, e.g., \cite{Li2012}.

Extensive work in cognitive science has shown that human behavior can be well modeled by objective-driven optimization \cite{Baker2007,Neumann2007,Luce2012}. In this context, a goal-based planning method is proposed in \cite{Ziebart2009} to predict future pedestrian trajectories. References \cite{Fisac2018} and \cite{Bajcsy2019} provide a Bayesian framework to reason about the model confidence. The authors of \cite{Hawkins2018} assume that humans are rational and try to control their actions to avoid collision. In \cite{Fisac2018}, the authors assume that the humans are irrational and it is the responsibility of the robot to maintain a safe distance from the human at all times. In \cite{Fisac2018}, the robot models the human as more likely to choose actions which minimize a goal objective function. We use a goal-based method to model the human's actions in this paper. However, we do not set any presumption on the rationality of the human. More precisely, we model the human's actions by means of a combination of two separate objective functions, a goal objective function and a safety objective function, where the robot learns this combination online. As will be seen later, this formulation allows us to add the capability of indirect communication between the robot and the human to understand the human's actual intention, and consequently reduce conservatism (i.e., improve efficiency of the robot).

Once a predictive human model is developed, the robot can use this model to generate a safe and efficient plan. Several planning schemes have been proposed in the literature. In \cite{Wilcox2012}, the authors introduced the adaptive preferences algorithm that computes a flexible optimal policy for robot scheduling and control in assembly manufacturing. In \cite{Ding2014}, a method has been proposed to optimize the task assignment such that the cycle time is shortened, and consequently the productivity is increased. Probabilistic wait-sensitive task planning have been proposed in \cite{Hawkins2013,Hawkins2014} to optimize the robot tasks with respect to the posterior human action distributions, reducing the total wait time of the human. In \cite{Tanaka2012,Kanazawa2019}, the authors proposed a motion planning scheme based on the human's trajectory prediction to improve efficiency. Genetic algorithms have also been utilized in planners, e.g., \cite{Baizid2015}. The notion of the virtual plane is used in \cite{Belkhouche2009} for path planning and navigation in dynamic environments. In \cite{Aoude2013}, the authors have developed a path planning framework to safely navigate robots, while avoiding dynamic obstacles with uncertain motion patterns. Finally, an optimal safe planner is proposed in  \cite{HosseinzadehICRA}, where the impacts of carelessness and boredom of humans have been taken into account. Our work builds upon the notion of danger awareness to create a scheme that can provide a safe and less-conservative plan, without degrading the efficiency of the robot.

%%%%%%%%%%%%%%%%%%%%%%%%%%%%%
\section{Problem Formulation}\label{sec:ProblemStatement}
\begin{figure}
\centering
\includegraphics[width=6cm]{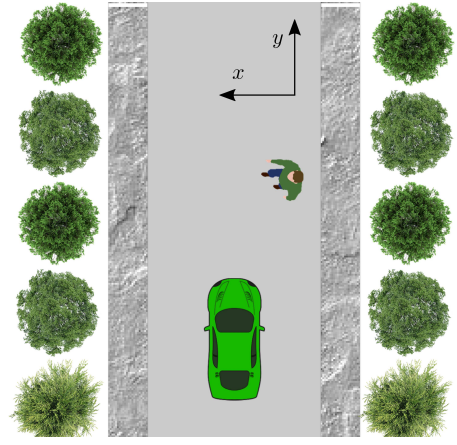}
\caption{Running example: the interaction between a self-driving car and a pedestrian.}
\label{Fig:Example}
\end{figure}

Consider a human-robot interaction, in which the robot and the human are moving to two different present goal locations. In the following, we formulate the problem for a general interaction, while the interaction between a self-driving car and a pedestrian will be being used as a running example to show the utility of the proposed method. The running example is demonstrated in Fig. \ref{Fig:Example}.

%%%%%%%%%%%%%%%%%%%%%%%%%%%%%%%
\subsection{Robot Model}
The robot can be modeled as 
\begin{align}\label{eq:robotmodel}
x_R[t+1]=f_R\left(x_R[t],u_R[t]\right),
\end{align}
where $x_R[t]\in\mathbb{R}^{n_R}$ and $u_R[t]\in\mathbb{R}^{m_R}$ are respectively the state and control action of the robot at time $t$, with $n_R$ and $m_R$ as the dimensions of the robot state-space and the robot control action, respectively.

Let $g_R\in\mathbb{R}^{n_R}$ be the goal state of the robot. Suppose that the robot control action should belong to $\mathcal{U}_R$ at all times, i.e., $u_R[t]\in\mathcal{U}_R,~\forall t\geq0$. We assume that the robot uses a receding predictive control to reach the state $g_R$, while avoiding collisions with the human. In particular, considering $Q_g^R(x_R[t],u_R[t],g_R):\mathbb{R}^{n_R}\times\mathbb{R}^{m_R}\times\mathbb{R}^{n_R}\rightarrow\mathbb{R}$ as the objective function corresponding to the goal state $g_R$, and $[t,t+T_R]$ with $T_R\in\mathbb{R}_{\geq0}$ as the prediction horizon, the robot solves the following optimization problem at time $t$ to compute the optimal control actions over the interval $[t,t+T_R]$:
\begin{align}\label{eq:RobotOP}
u^\ast_R[t:t+T_R],d_R=\left\{
\begin{array}{rl}
     & \arg\min\limits_{u_R[k]}\sum_{k=t}^{t+T_R}Q_g^R(\cdot) \\
  \text{s.t.}   & u_R[k]\in\mathcal{U}_R,~\forall k\\
  & \text{The model given in \eqref{eq:robotmodel}} \\
  & P_{Coll}[k]\leq P_{th},~\forall k
\end{array}
\right.,
\end{align}
where $k\in\{t,\cdots,t+T_R\}$ and $u^\ast_R[t:t+T_R]=\left[\begin{matrix}u_R^{\ast^\top}[t] & \cdots & u_R^{\ast^\top}[t+T_R]\end{matrix}\right]^\top$, but only implements the action $u_R^\ast[t]$, and then solves \eqref{eq:RobotOP} again at the next time instance, repeatedly. In \eqref{eq:RobotOP}, $d_R$ is the on/off status of the danger signal (will be discussed in Section \ref{sec:DangerNotifier}), $P_{Coll}[t]\in[0,1]$ is the the probability of collision between the human and robot at time $t$ (will be discussed later in Section \ref{sec:collision}), and $P_{th}\in[0,1]$ is the threshold value.%, and \textcolor{blue}{$\mathcal{X}\subseteq\mathbb{R}^{n_R}$ is a \textit{robust control invariant set}. The last constraint ensures \textit{recursive feasibility} of \eqref{eq:RobotOP} (see Section \ref{sec:InvariantSet})}.

\paragraph*{Running Example}
The state of the self-driving car is $x_R[t]=\left[\begin{matrix}p_R^x[t] & p_R^y[t]\end{matrix}\right]^\top\in\mathbb{R}^2$, where $p_R^x[t]$ and $p_R^y[t]$ are x and y positions of the self-driving car at time $t$, respectively. The control action of the self-driving car is $u_R[t]=\left[\begin{matrix}u_R^x[t] & u_R^y[t]\end{matrix}\right]^\top\in\mathbb{R}^2$, where $u_R^x[t]$ and $u_R^y[t]$ are the directional velocities along x and y axes at time $t$, respectively. The dynamical model of the self-driving car is then $x_R[t+1]=x_R[t]+u_R[t]$. The target position for the self-driving car is $g_R=\left[\begin{matrix}g_R^x & y_R^y\end{matrix}\right]^\top\in\mathbb{R}^2$. For the sake of simplicity, we assume that the self-driving car can only move along the y axis, i.e., $p_R^x[t]=g_R^x,~u_R^x[t]=0,~\forall t\geq0$. Suppose that $\mathcal{U}_R=\left\{\left[\begin{matrix}0~0\end{matrix}\right]^\top,\left[\begin{matrix}0~v_R/2\end{matrix}\right]^\top,\left[\begin{matrix}0~v_R\end{matrix}\right]^\top\right\}$, where $v_R$ is the basic velocity of the self-driving car. This means that the self-driving car can move at either zero speed, or half speed, or full speed. Finally, the goal objective function of the self-driving car is $Q_g^R(x_R[t],u_R[t],g_R)=\theta_1\left(p_R^y[t]+u_R^y[t]-g_R^y\right)^2+\theta_2\left(u_R^y[t]\right)^2$, where $\theta_1\in\mathbb{R}_{>0}$ and $\theta_2\in\mathbb{R}_{\geq0}$ are design parameters. In this objective function, the first term is the Euclidean distance between the target position and the self-driving car at time $t$, and the second term is the distance traveled by the self-driving car within one time step.

%%%%%%%%%%%%%%%%%%%%%%%%%%%%%%%
\subsection{Human Model}\label{sec:HumanModel}
The human can be modeled as
\begin{align}\label{eq:humanmodel}
x_H[t+1]=f_H\left(x_H[t],u_H[t]\right),
\end{align}
where $x_H[t]\in\mathbb{R}^{n_H}$ and $u_H[t]\in\mathbb{R}^{m_H}$ are respectively the human's state and action at time $t$, with $n_H$ as the dimension of the human state-space and $m_H$ as the dimension of the human's action.

Let $g_H\in\mathbb{R}^{n_H}$ be the human's goal state. Suppose that the human's action should belong to $\mathcal{U}_H$ at all times, i.e., $u_H[t]\in\mathcal{U}_H,~\forall t\geq0$.

As discussed in \cite{Luce2005,Baker2007,Fisac2018}, human's action can be well modeled by objective-driven optimization. In the considered HRI, we can model the human's action as optimizing a combination of a goal objective function and a safety objective function. In mathematical terms, the human's action at time $t$ is the solution of the following optimization problem:
\begin{align}\label{eq:HumanOP}
u_H^\ast[t]=\left\{
\begin{array}{rl}
     & \arg\min\limits_{u_H}Q_g^H\left(x_H[t],u_H,g_H\right)\\
     & ~~~~~~~~~~~+\beta Q_s^H\left(x_H[t],u_H,\hat{x}_R[t]\right) \\
  \text{s.t.}   & \text{The model given in \eqref{eq:humanmodel}} \\
  & u_H\in\mathcal{U}_H
\end{array}
\right.,
\end{align}
where $Q_g^H(x_H[t],u_H[t],g_H):\mathbb{R}^{n_H}\times\mathbb{R}^{m_H}\times\mathbb{R}^{n_H}\rightarrow\mathbb{R}$ is the goal objective function corresponding to the goal state $g_H$, and $Q_s^H(x_H[t],u_H[t],\hat{x}_R[t]):\mathbb{R}^{n_H}\times\mathbb{R}^{m_H}\times\mathbb{R}^{n_R}\rightarrow\mathbb{R}$ is the safety objective function. In \eqref{eq:HumanOP}, $\hat{x}_R[t]\in\mathbb{R}^{n_R}$ is an estimation of the state of the robot at time $t$ computed by the human. Most importantly, $\beta$ is a binary variable (i.e., $\beta\in\{0,1\}$) that we refer to as the danger awareness coefficient. A common interpretation of $\beta=0$ is a human who does not see the robot, is careless, or presumes that it is the responsibility of the robot to keep a safe distance. Whereas, $\beta=1$ means that the human is aware of the danger, is risk-averse, and acts properly to reduce the risk.

\begin{remark}
The estimation $\hat{x}_R[t]$ can be modeled as $\hat{x}_R[t]=x_R[t]+\epsilon[t]$, with $\epsilon[t]\in\mathbb{R}^{n_R}$ as a zero-mean Gaussian random variable with covariane $\Sigma\in\mathbb{R}_{\geq0}^{n_R\times n_R}$, i.e., $\epsilon[t]\sim\mathcal{N}(0,\Sigma)$.
\end{remark}

\paragraph*{Running Example}
The pedestrian is an adult, whose state is $x_H[t]=\left[\begin{matrix}p_H^x[t] & p_H^y[t]\end{matrix}\right]^\top\in\mathbb{R}^2$, where $p_H^x[t]$ and $p_H^y[t]$ are x and y positions of the pedestrian at time $t$, respectively. The pedestrian's action is $u_H[t]=\left[\begin{matrix}u_H^x[t] & u_H^y[t]\end{matrix}\right]^\top\in\mathbb{R}^2$, where $u_H^x[t]$ and $u_H^y[t]$ are the directional velocities along x and y axes at time $t$, respectively. The pedestrian's model is then $x_H[t+1]=x_H[t]+u_H[t]$. The target position for the pedestrian is $g_H=\left[\begin{matrix}g_H^x & g_H^y\end{matrix}\right]^\top\in\mathbb{R}^2$. For the sake of simplicity, we assume that the pedestrian can only move along the x axis, i.e., $p_H^y[t]=g_H^y,~u_H^y[t]=0,~\forall t\geq0$.  Suppose that $\mathcal{U}_H=\left\{\left[\begin{matrix}-2v_H~0\end{matrix}\right]^\top,\left[\begin{matrix}-v_H~0\end{matrix}\right]^\top,\left[\begin{matrix}0~0\end{matrix}\right]^\top,\left[\begin{matrix}v_H~0\end{matrix}\right]^\top,\left[\begin{matrix}2v_H~0\end{matrix}\right]^\top\right\}$, where $v_H$ is the pedestrian's basic walking velocity. This means that the pedestrian can stop, walk, or run in either directions. The pedestrian's goal objective function is $Q_g^H(x_H[t],u_H[t],g_H)=\theta_3\left( p_H^x[t]+u_H^x[t]-g_H^x\right)^2+\theta_4\left(u_H^x[t]-v_H\right)^2$, where $\theta_3,\theta_4\in\mathbb{R}_{>0}$ are design parameters. In this objective function, the first term is the Euclidean distance between the target position and the pedestrian at time $t$, and the second term ensures that the pedestrian walks toward the target position. Finally, the safety objective function is $Q_s^H\left(x_H[t],u_H,\hat{x}_R[t]\right)=\theta_5e^{-\theta_6\cdot\text{dist}[t]}$, where $\theta_5,\theta_6\in\mathbb{R}_{>0}$ are design parameters, and $\text{dist}[t]\in\mathbb{R}_{\geq0}$ is the distance between self-driving car and the pedestrian estimated by the pedestrian at time $t$. This safety objective function is indeed a penalty function on the distance between the pedestrian and the self-driving car, such that larger distance produces lower penalty value. This formulation is plausible, as humans continue walking toward the target position when there is large distance.

%%%%%%%%%%%%%%%%%%%%%%%%%%%%%%%
\subsection{Danger Signaling System}\label{sec:DangerNotifier}
Despite majority of work in the literature, we assume that the humans can be influenced by the actions of the robots. Indeed, we believe that assuming that humans irrationally operate in human-robot environments and intentionally ignore the robots not only is unrealistic, but also severely harms the efficiency of the robots.

To address this issue, we assume that the robot is equipped with a proper pre-collision method which uses signals/indicators to alert the danger to the human (e.g., visual indicators \cite{Baraka2016} and auditory signals \cite{Wogalter2018}). As discussed in Section \ref{sec:introduction}, the main goal of employing the danger signaling is to improve the robot's ability to estimate the state of the human danger awareness, and to maintain the efficiency of the robot in reaching the goal state without being influenced by the human's possible unsafe actions. We denote the on/off status of the danger signal by the binary variable $d_R$, where $d_R=0$ if the signal is off and $d_R=1$ if it is on. The robot switches the signal on when the constraint $P_{Coll}[k]\leq P_{th}$ in \eqref{eq:RobotOP} is active for any $k\in\{t,\cdots,t+T_R\}$ (i.e., this constraint effects the obtained results).

\begin{remark}\label{remark:dangernotifier}
From a technical viewpoint, the danger signaling can actively perturb the environment so to improve the efficiency and safety of the interaction by: 1) acquainting an unaware human or a human who underestimates the danger (i.e., changing the value of $\beta$ from 0 to 1), and 2) helping the human to reduce the estimation error $\epsilon[t]$ (e.g., the error may be large due to bright sun glare in the human's eyes).
\end{remark}

\paragraph*{Running Example}
The self-driving car uses the high beams to notify the possible collision between the car and the pedestrian.

%%%%%%%%%%%%%%%%%%%%%%%%%%%%%%%
\subsection{Problem Statement}
At this stage we define the following problem. 

\begin{problem}\label{problem}
Consider a HRI, in which the robot and the human are moving to two different goal locations. Suppose that the robot model is as in \eqref{eq:robotmodel}, and the robot uses the receding predictive control given in \eqref{eq:RobotOP} to determine the next action. Suppose that the human's model is as in \eqref{eq:humanmodel}, and the human decides the next action via optimization problem \eqref{eq:HumanOP}. Suppose that the robot uses the danger signaling system to alert the danger to the human. Provide a method to ensure that both agents reach their goal states, while the safety of the human and the robot is guaranteed. 
\end{problem}

\begin{figure}[!t]
\centering
\includegraphics[width=8.5cm]{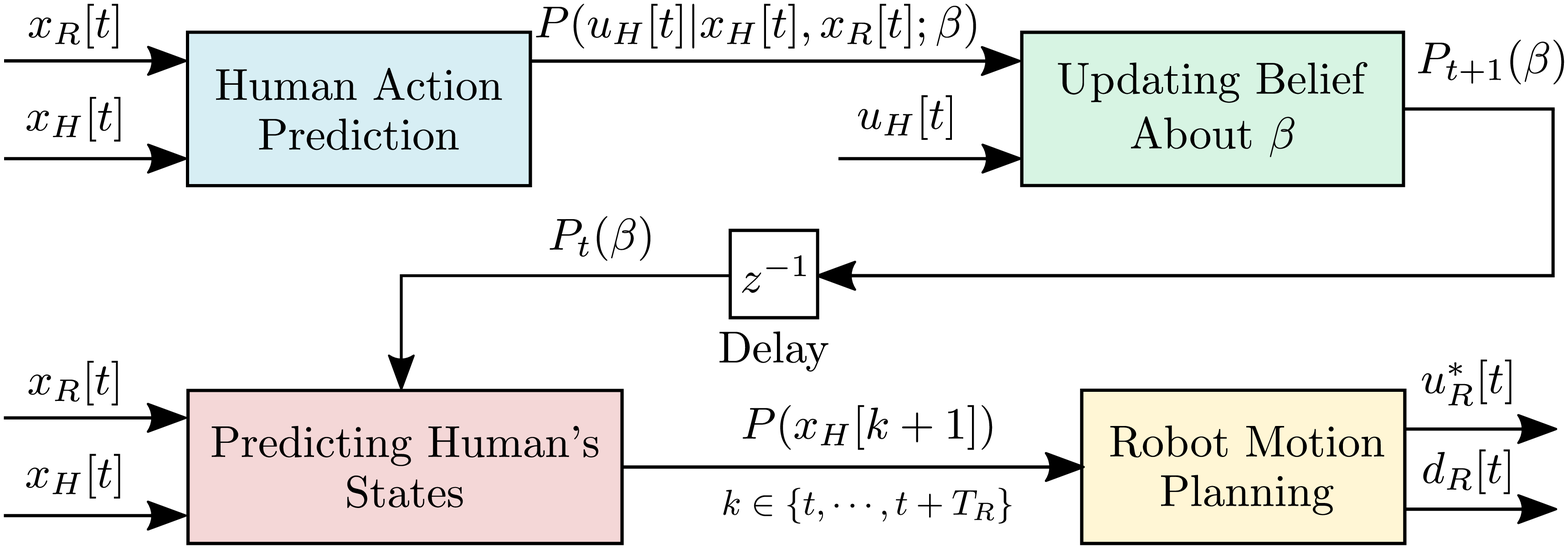}
\caption{The general structure of the proposed planning scheme.}
\label{fig:generalstructure}
\end{figure}

In order to solve Problem \ref{problem}, 
assuming that the robot can observe human's location and action, we will develop a scheme to provide probabilistically safe robot actions to guide the robot to the goal state, without any collision with the human. The structure of the proposed scheme is depicted in Fig.  \ref{fig:generalstructure}. The gist of this scheme is the development of a predictive model of the human's motion, whose values are computed through posterior calculations based upon observations performed by the robot. The proposed scheme will be discussed in detail in the following sections.

%%%%%%%%%%%%%%%%%%%%%%
\section{Predictive Human Model}\label{sec:PredictiveHumanModel}
According to \eqref{eq:RobotOP}, the robot determines its actions by taking into account the future states of the human. This mean that the robot must employ a predictive human model to predict human's states in the future. The accuracy of these predictions and the method used to plan around them determine the safety of the interaction.

According to \eqref{eq:HumanOP}, the human decides next action according to the objective functions $Q_g^H(\cdot)$ and $Q_s^H(\cdot)$, and the danger awareness coefficient $\beta$. We assume that the robot knows the objective functions $Q_g^H(\cdot)$ and $Q_s^H(\cdot)$. This assumption is reasonable, as the robot can learn these functions from prior human's motions or these functions can be explicitly provided by the system designers. However, any presumption on the value of the coefficient $\beta$ will often be wrong in practice; humans may have different opinions of safety, or they may be less attentive than normal when facing robots presuming that it is the responsibility of the robot to maintain a safe distance. Thus, the robot must be able to timely reason over the value of $\beta$ to produce a reliable distribution of human's states in the future.

In what follow, first, we will propose a probabilistic Boltzmann model to predict the human's action. Then, an update method will be proposed to update the belief on the value of the danger awareness coefficient. Finally, it will be discussed how the robot can predict the probability of a collision in the future.

\begin{figure*}[!b]
\hrule
\setcounter{equation}{9}
\begin{align}\label{eq:collisonProbability}
P_{Coll}[k]=&P\big(x_R[k]\in\pi(x_H[k]),x_R[k-1]\not\in\pi(x_H[k-1]),\cdots,x_R[t+1]\not\in\pi(x_H[t+1])\big)\nonumber\\
=&P\big(x_R[k]\in\pi(x_H[k])|x_R[k-1]\not\in\pi(x_H[k-1]),\cdots,x_R[t+1]\not\in\pi(x_H[t+1])\big)\nonumber\\
&\times P\big(x_R[k-1]\not\in\pi(x_H[k-1])|x_R[k-2]\not\in\pi(x_H[k-2]),\cdots,x_R[t+1]\not\in\pi(x_H[t+1])\big)\nonumber\\
&\times\cdots\times P\big(x_R[t+1]\not\in\pi(x_H[t+1])\big) \nonumber\\
=&P\big(x_R[k]\in\pi(x_H[k])|x_R[k-1]\not\in\pi(x_H[k-1]),\cdots,x_R[t+1]\not\in\pi(x_H[t+1])\big)\nonumber\\
&\times\bigg(1-P\big(x_R[k-1]\in\pi(x_H[k-1])|x_R[k-2]\not\in\pi(x_H[k-2]),\cdots,x_R[t+1]\not\in\pi(x_H[t+1])\big)\bigg)\nonumber\\
&\times\cdots\times\bigg(1-P\big(x_R[t+1]\in\pi(x_H[t+1])\big)\bigg).
\end{align}
\end{figure*}

\subsection{Human Action Prediction}\label{sec:HumanActionPrediction}
As mentioned above, we assume that the robot knows the objective functions $Q_g^H(\cdot)$ and $Q_s^H(\cdot)$. Under this assumption, the robot can predict the human’s action as a probability distribution over actions conditioned on the human's state.

\paragraph*{Running Example}
The formulation of the goal objective function $Q_g^H(\cdot)$ is quite straightforward, as the pedestrian wants to walk from one sidewalk to the other sidewalk of the street. For what concerns the safety objective function $Q_s^H(\cdot)$, it is possible to learn it for pedestrians\footnote{We understand that there might be some differences between people. However, this paper does not aim to deal with the differences.} based upon behavioral patterns \cite{Zacharias2001,Campbell2003,Mehta2008,McAslan2017}. Note that since $\beta$ multiplies $Q_s^H(\cdot)$ in \eqref{eq:HumanOP}, either $\beta=0$ (i.e., unaware humans) or $Q_s^H(\cdot)=0$ (e.g., children who do not recognize the danger) has the same results in human actions.

The robot uses the following mixture distribution to model the human's behavior:\setcounter{equation}{4}
\begin{align}\label{eq:predictionFinal}
P(u_H|x_H,x_R;\beta)=&(1-\omega_H)\cdot P_d(u_H|x_H,x_R;\beta)\nonumber\\
&+\omega_H\cdot P_r(u_H),
\end{align}
where $P_d(u_H|x_H,x_R;\beta)$ models the human's deliberate behavior, $P_r(u_H)$ models the human's random behavior, and $\omega_H\in[0,1]$ is the mixture weight. Note that \eqref{eq:predictionFinal} should be computed for every $u_H\in\mathcal{U}_H$ and every $\beta\in\{0,1\}$.

The function $P_d(u_H|x_H,x_R;\beta)$ describes the probability distribution of the human's action if the human chooses the action according to the the goal and safety objective functions. Assuming that the robot can observe the human's state, one possible way to model the human's deliberate behavior is to use the Boltzmann distribution, as follows:
\begin{align}\label{eq:prediction}
P_d(u_H|x_H,x_R;\beta)\propto e^{-\gamma\left(Q_g^H\left(x_H,u_H,g_H\right)+\beta Q_s^H\left(x_H,u_H,x_R\right)\right)},
\end{align}
where $\gamma\gg1$. Note that \eqref{eq:prediction} should be computed for every $u_H\in\mathcal{U}_H$ and every $\beta\in\{0,1\}$.

\begin{remark}
Selecting a large $\gamma$ ensures that model \eqref{eq:prediction} treats the human as more likely to choose the action that minimizes the cost function given in \eqref{eq:HumanOP}. More precisely, $P_d(u_H^\ast[t]|x_H[t],x_R[t];\beta)\approx1$, where $u_H^\ast[t]$ is as in \eqref{eq:HumanOP}, and $P_d(u_H[t]|x_H[t],x_R[t];\beta)\approx0$, where $u_H[t]\neq u_H^\ast[t]$. 
\end{remark}

\begin{remark}
The robot uses the actual state $x_R[t]$ in \eqref{eq:prediction} to predict the human's actions, while the human uses the estimation $\hat{x}_R[t]$ in \eqref{eq:HumanOP} to decide the next action. Thus, if the estimation error $\epsilon[t]$ is large, we may have $P_d(u_H^\ast[t]|x_H[t],x_R[t];\beta)\not\approx1$, where $u_H^\ast[t]$ is as in \eqref{eq:HumanOP}. \end{remark}

In reality, the human may choose a random action by completely ignoring the objective functions for any reason. The robot makes use of a uniform distribution to model the human's random behavior, as follows
\begin{align}
P_r(u_H)=\frac{1}{\vert\mathcal{U}_H\vert},~\forall u_H\in\mathcal{U}_H,
\end{align}
where $\vert\mathcal{U}_H\vert$ is the cardinality of the set $\mathcal{U}_H$. Note that this uniform distribution can be given a more practical interpretation related to the accuracy of the estimation $\hat{x}_R[t]$ and/or the reliability of the detectors used by the robot to observe the human's state and action.

\paragraph*{Running Example} The uniform distribution $P_r(u_H)$ models cases in which the pedestrian stops on the road or walks in the opposite direction of the target position $g_H$ for any reason other than safety.

\subsection{Real-Time Update of the Belief About the Coefficient $\beta$}\label{sec:UpdateBelief}
The danger awareness coefficient $\beta$ can be seen as a hidden state. Given \textit{a priori} belief about the danger awareness coefficient (i.e., $P_0(\beta),~\forall\beta$), the robot can use the observations to update the belief about the danger awareness coefficient by applying the Bayes' rule. In mathematical terms, by observing the human's state and action at time $t$, the robot can update the belief about the danger awareness coefficient via the following Bayesian update:
\begin{align}\label{eq:updatebelief}
P_{t+1}(\beta)=\frac{P(u_H[t]|x_H[t],x_R[t];\beta)P_t(\beta)}{\sum_{\bar{\beta}}P(u_H[t]|x_H[t],x_R[t];\bar{\beta})P_t(\bar{\beta})},~\forall\beta,
\end{align}
where $P(u_H[t]|x_H[t];\beta)$ is as in \eqref{eq:predictionFinal}. Note that since the set of $\beta$ values and the set of human's possible actions $\mathcal{U}_H$ are small, update rule \eqref{eq:updatebelief} can be implemented in real-time. It is noteworthy that $P_t(\beta=1)$ is the robot's belief at time $t$ about the likelihood that the human is aware of the danger. %The danger awareness coefficient estimated by the robot at time $t$ can then be computed as
%\begin{align}
%\hat{\beta}[t]=\arg\;\max_{\beta\in\{0,1\}}P_{t}(\beta).
%\end{align}

\subsection{Computation of the Probability of Collision}\label{sec:collision}
Suppose that the human's state-space is divided into $N_c$ discrete grid cells. Thus, by observing $x_H[t]$, the probability of the human's state in the time interval $[t+1,t+T_R]$ can be predicted via the following recursive update rule:
\begin{align}\label{eq:stateprediction}
P\left(x_H[k+1]\right)\propto&\sum_{x_H[k],u_H[k]}\sum_{\beta}P\left(x_H[k+1]|x_H[k],u_H[k]\right)\nonumber\\
&\cdot P\left(u_H[k]|x_H[k],x_R[k];\beta\right)\cdot P_t(\beta),
\end{align}
for $k=t,\cdots,t+T_R-1$, where $P\left(u_H[k]|x_H[k],x_R[k];\beta\right)$ can be computed via \eqref{eq:predictionFinal}, $P_t(\beta)$ can be computed via \eqref{eq:updatebelief}, and $P\left(x_H[k+1]|x_H[k],u_H[k]\right)$ is equal to 1 if $x_H[k+1]$, $x_H[k]$, and $u_H[k]$ satisfy \eqref{eq:humanmodel}, and is equal to zero otherwise.

Note that \eqref{eq:stateprediction} should be computed for every grid cell in the human's state-space. In other words, for each $k$, the likelihood of the human's state being in all $N_c$ cells should be computed. Also, note that we use $P_t(\beta)$ in \eqref{eq:stateprediction} to predict the human's state, meaning that the danger awareness coefficient is assumed to be constant within the prediction horizon. Due to the receding nature of the planner given in \eqref{eq:RobotOP}, this assumption does not hamper the performance of the planner.

Once the probability distribution of the human's state over the time interval $[t+1,t+T_R]$ is generated, the robot should computed the probability of collision $P_{Coll}[k],~k=t+1,\cdots,t+T_R$. Note that $P_{Coll}[t]=0$ in \eqref{eq:RobotOP}, as there is no collision at the current time (i.e., time $t$).

Let $\pi(x_H[k]),~k=t+1,\cdots,t+T_R$ be a neighborhood around the predicted human's state $x_H[k]$. This neighborhood should be determined according to the predefined safe distancing measures, the effect of possible modeling and tracking errors on the predictions, and the effect of gridding the human's state-space (i.e., quantization error).

By construction, the probability of a collision event at time $k$ (for $k\in\{t+1,\cdots,t+T_R\}$) can be computed \cite{Fisac2018} as the probability that $x_R[k]$ is inside the neighborhood $\pi(x_H[k])$, without any collisions prior to $k$. This probability is presented in \eqref{eq:collisonProbability}. Note that \eqref{eq:collisonProbability} should be computed recursively.

\begin{remark}
Since the effect of gridding the human's state-space is reflected in the function $\pi(\cdot)$, safety analyses of this paper are valid even with a small $N_c$ (i.e., large cells).
\end{remark}

\begin{remark}
According to \eqref{eq:collisonProbability}, the probability of collision for $k\in\{t+1,t+T_R\}$ can be upper bounded as
\setcounter{equation}{10}
\begin{align}\label{eq:collisonProbability2}
P_{Coll}[k]\leq P\left(x_R[k]\in\pi(x_H[k])\right),
\end{align}
which means that collision probabilities over time are independent. This upper bound may lead to a conservative solution. However, it can significantly reduce the computational complexity of optimization problem \eqref{eq:RobotOP}.
\end{remark}

The following proposition elucidates how \eqref{eq:collisonProbability} (or \eqref{eq:collisonProbability2}) can be computed according to \eqref{eq:stateprediction}.

\begin{proposition}
Suppose that $x_R[k],~k\in\{t+1,\cdots,t+T_R\}$ is the robot trajectory within the prediction horizon. Then, $P\left(x_R[k]\in\pi(\alpha)\right)$ where $\alpha$ is a cell in the human's state-space is equal to the probability that the human's state at time $k$ is $\alpha$, i.e., $P(x_H[k]=\alpha)$, which can be computed through \eqref{eq:stateprediction}.
\end{proposition}

\paragraph*{Running Example} We define $\pi(x_H[k])$ as a circle centered on $x_H[k]$ with radius $\rho$. Thus, $x_R[k]\in\pi(x_H[k])$ iff  $(p_H^x[k]-p_R^x[k])^2+(p_H^y[k]-p_R^y[k])^2\leq\rho^2$.

\begin{remark}
Setting $P_{th}=0$ provides a \textit{deterministically safe} trajectory. Such trajectories are usually very conservative, as they take into account the worst-case scenario. The obtained trajectory for $P_{th}\in(0,1]$ is probabilistically safe, which, in general, does not guarantee recursive feasibility. This is an issue in many planning schemes which are developed based upon a probabilistic human model, e.g., \cite{Aoude2013,Fisac2018,Bajcsy2019,Chapman2019,Keil2020}. Indeed, since the robot actions are limited and due to the imperfect human model, the recursive feasibility is unsurprisingly very difficult to satisfy without rendering the solution conservative. %One intuitive way to provide a strong safety certificate is to enlarge the neighborhood $\pi(\cdot)$. This approach  the robot is enforced to keep a large distance with the human; however, it renders the solution conservative. 
\end{remark}

\begin{figure*}[!t]
\centering
\hspace{-1cm}\includegraphics[width=5cm]{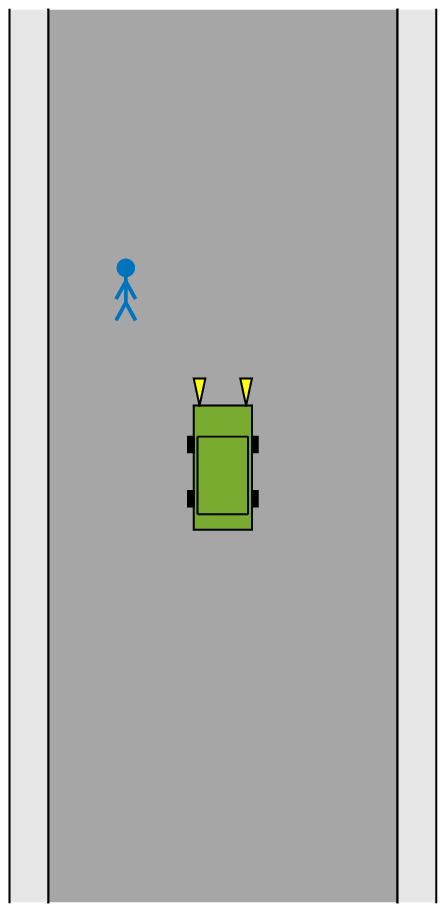}\hspace{-0.7cm}\includegraphics[width=4.1cm]{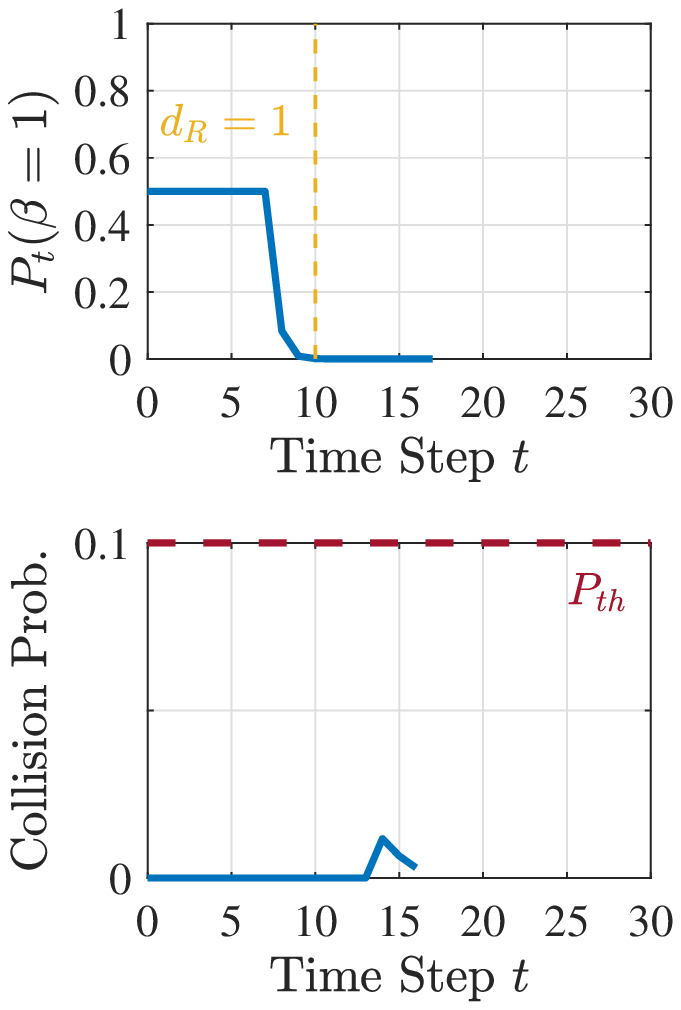}\hspace{-0.7cm}\includegraphics[width=5.7cm]{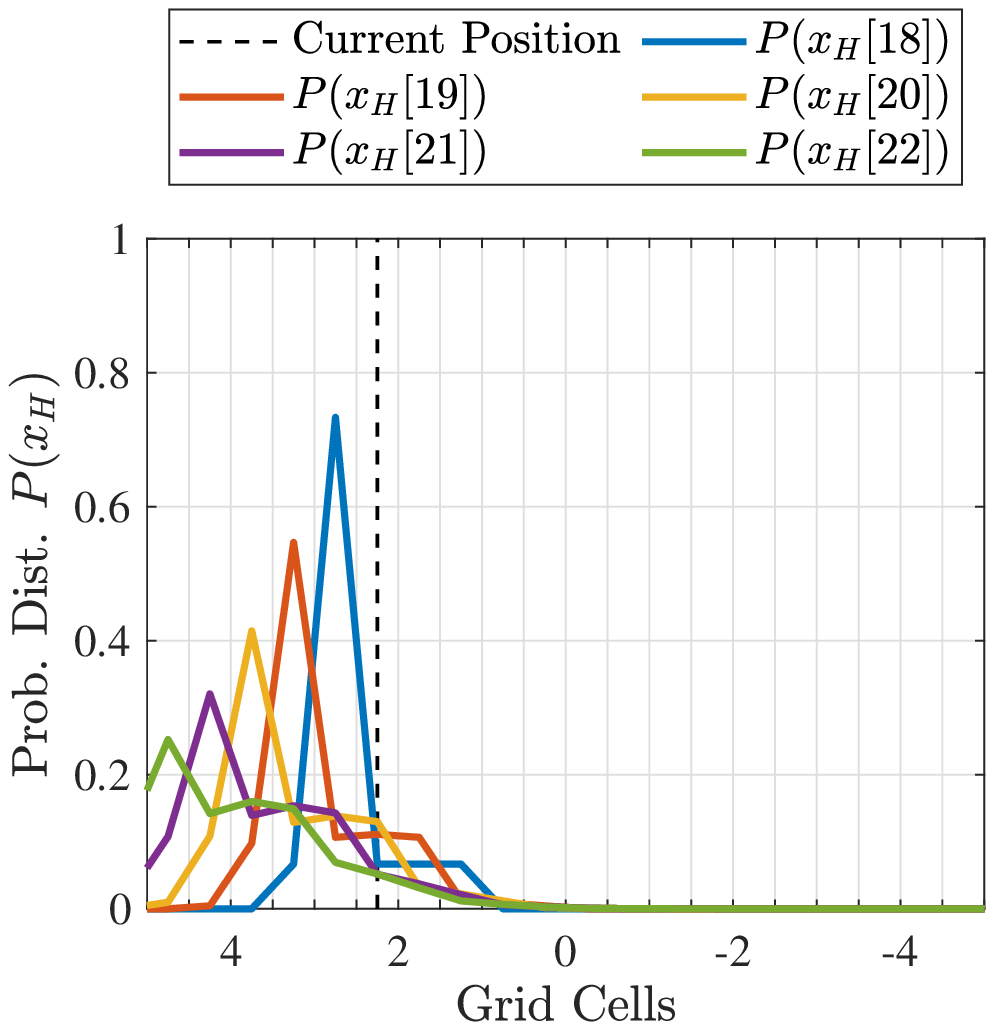}\hspace{-0.8cm}\includegraphics[width=5.6cm]{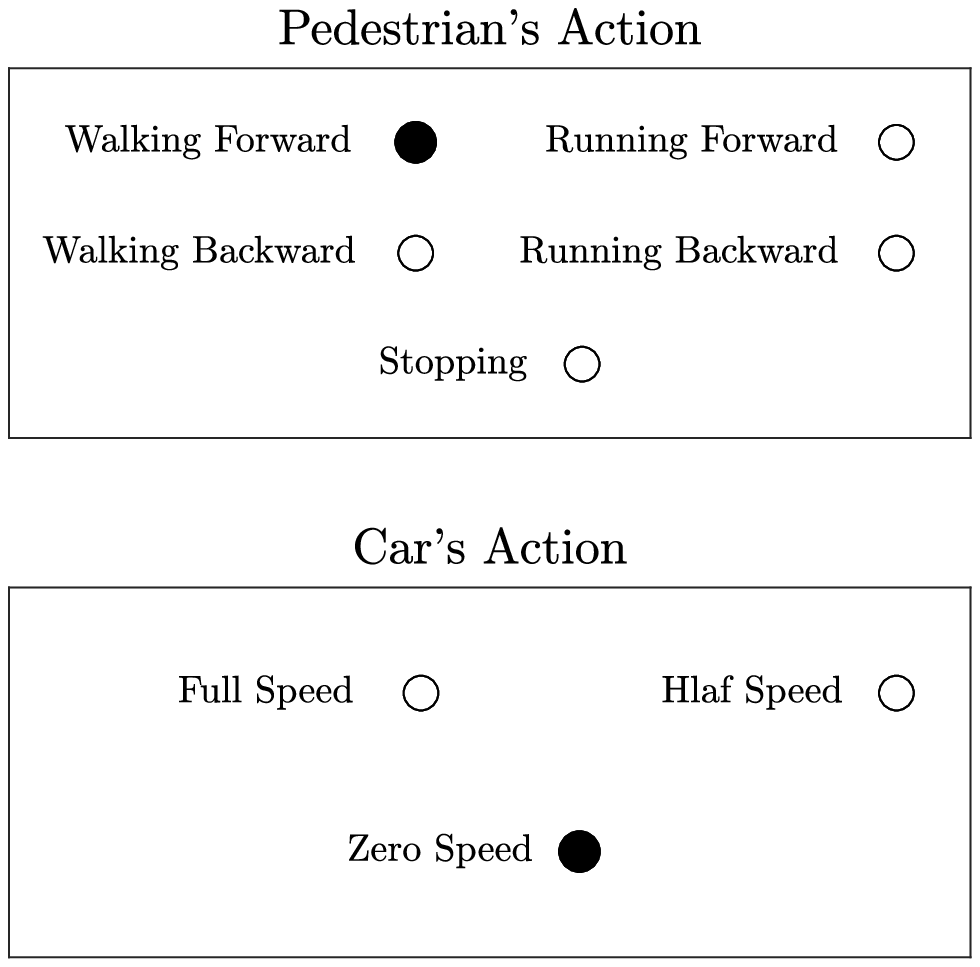}
\caption{A screenshot of the generated simulator shown in the accompanied video (see \url{https://youtu.be/_9UjDvZYT2U}). Left figure: the interaction between a self-driving car and a pedestrian in the street; the pedestrian moves from right to left, and the car moves from south to north. Middle-left figures: the top figure shows the time profile of the robot's belief about the likelihood that the human is aware of the danger, i.e., $P_t(\beta=1)$, and the bottom figure shows the probability of collision over the prediction horizon. Middle-right figure: probability distribution of the human's position in the street over the prediction horizon, computed at each time instant (here at time $t=17$); note that the pedestrian moves from right to left. Right figures: the top and bottom figures indicate the pedestrian and car actions at the current time instant, respectively.}
\label{fig:Simulator}
\hrule
\end{figure*}

%%%%%%%%%%%%%%%%%%%%%%
\section{The Proposed Planning Scheme}\label{sec:framework}
%Thus far, we have shown how to predict the human's next actions and states, how to compute probability of collision in the future, and how to update the belief about the human's opinion of cooperative safety. In this section, we will show how to build a planning scheme, based upon discussions in previous sections, to safely navigate the robot to the goal state. 

The general structure of the proposed planning scheme is shown in Fig. \ref{fig:generalstructure}. In this scheme, first, the robot observes the human's state $x_H[t]$ and its own state $x_R[t]$. Based on these observations, the robot generates two probability distributions: 1) a distribution over the human's next action, and 2) a distribution over the human's states in a future time interval. The former probability distribution will be used to update the belief on the human's risk-aversion level. The latter probability distribution will be used to compute probability of collision in a future time interval. The robot then solves an optimization problem to determine the next action and the on/off status of the danger signal. Finally, Once the human takes an action, the robot observes the action and updates the belief about the danger awareness coefficient. The proposed scheme can be summarized as Algorithm \ref{algorithm:algorithm}.

\begin{algorithm}[!t]
\caption{Planning Scheme}
\label{algorithm:algorithm}
\begin{algorithmic}[1]
\STATE Observe the human's state $x_H[t]$ and the state of the robot $x_R[t]$.
\STATE Compute the mixture distribution $P(u_H|x_H[t],x_R[t];\beta)$ for every $u_H\in\mathcal{U}_H$ and $\beta$ via \eqref{eq:predictionFinal}.
\STATE Compute the probability distribution of human's states $P(x_H[k])$ for $k\in\{t+1,t+T_R\}$ via \eqref{eq:stateprediction}. 
\STATE Compute the probability of collision $P_{Coll}[k]$ for $k\in\{t+1,t+T_R\}$ via \eqref{eq:collisonProbability} (or \eqref{eq:collisonProbability2}).
\STATE Determine the action of the robot $u_R^\ast[t]$ and the on/off status of the danger signal $d_R$ via \eqref{eq:RobotOP}. 
\STATE Observe the human's action $u_H[t]$.
\STATE Update the belief about the danger awareness coefficient via \eqref{eq:updatebelief}, i.e., compute $P_{t+1}(\beta)$ for all $\beta$.
\end{algorithmic}
\end{algorithm}

%%%%%%%%%%%%%%%%%%%%%%%%%%%%%%%%%%%%%%%%%
\section{Simulation Study}\label{sec:simulation}
In order to demonstrate the effectiveness of the proposed scheme, we simulate the considered running example, i.e., the interaction between a self-driving car and a pedestrian shown in Fig. \ref{Fig:Example}. We assume that $g_R=[0~80]^\top$, $g_H=[5~10]^\top$, $v_R=2$, $v_H=0.5$, $\omega_H=0.1$, $P_{th}=0.1$, $T_R=5$, $\gamma=1000$, $\theta_1=1$, $\theta_2=0.5$, $\theta_3=2.5$, $\theta_4=8\times10^{-3}$, $\theta_5=300$, $\theta_6=6\times10^{-3}$, $\Sigma=1$, and $P_0(\beta=0)=P_0(\beta=1)=1/2$. The simulations are carried out using MATLAB/Simulink package, on Intel(R) Core(TM)i7-7500U CPU 2.70 GHz with 16.00 GB of RAM. We use the \texttt{YALMIP} toolbox \cite{yalmip} to solve the optimization problems.

In order to have a visual demonstration of the considered interaction between a pedestrian and a self-driving car, a simulator has been generated. Fig. \ref{fig:Simulator} presents an overview of the generated simulator. A video of operation of the simulator is available at the URL: \url{https://youtu.be/_9UjDvZYT2U}.

\subsection{Impact of the Danger Signaling System}
\begin{figure}[!t]
\centering
\includegraphics[width=7cm]{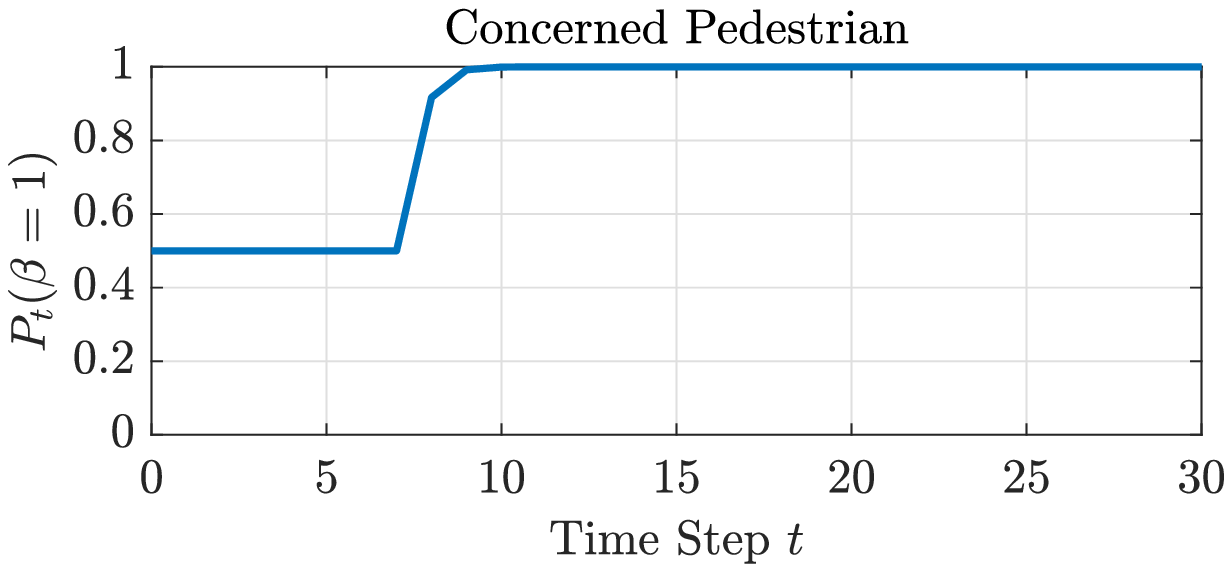}\\
\includegraphics[width=7cm]{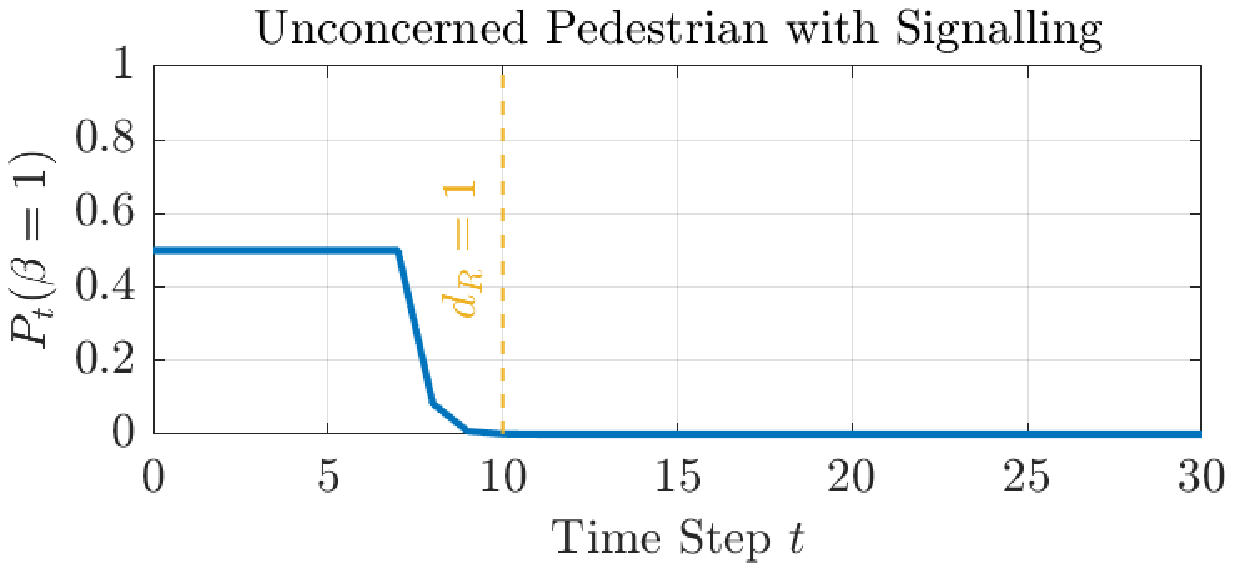}\\
\includegraphics[width=7cm]{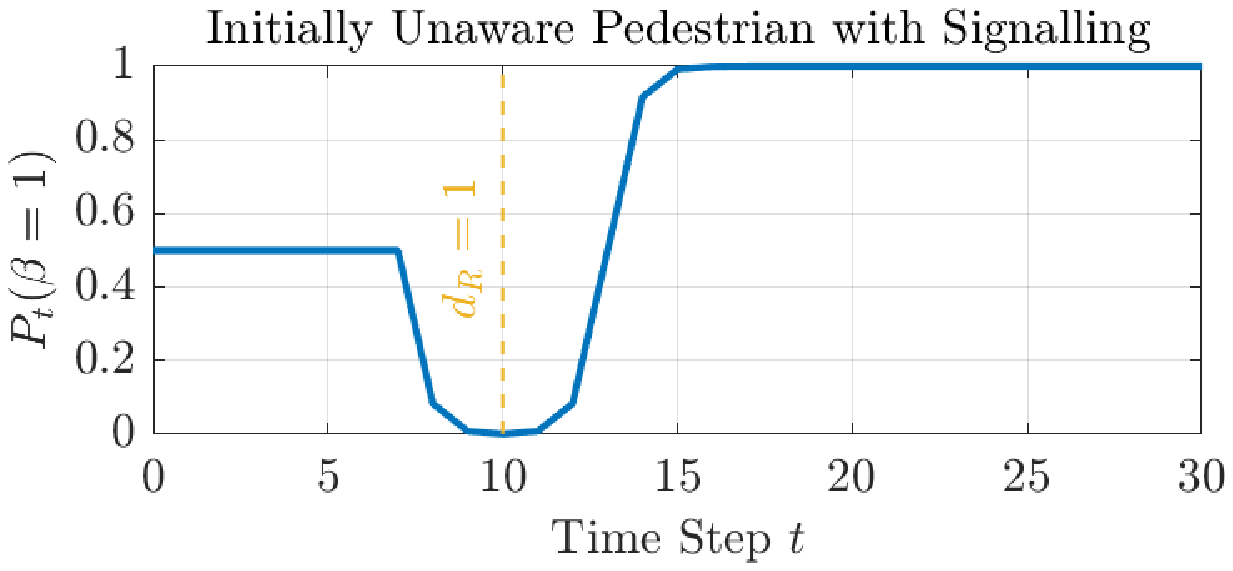}
\caption{Time profile of the robot's belief about the likelihood that the pedestrian is aware of the danger. Top figure: the pedestrian is concerned, i.e., is aware of the danger and engages in the safety enforcement. Middle figure: the pedestrian is unconcerned, i.e., either is unaware of the danger or does not care. Bottom figure: the danger signaling system acquaints the concerned but unaware pedestrian.}
\label{fig:BeliefProfile}
\end{figure}

As discussed in Remark \ref{remark:dangernotifier}, the danger signaling system can improve the efficiency and safety by acquainting the unaware pedestrian. This aspect is shown in Fig. \ref{fig:BeliefProfile}. As seen in this figure, when the pedestrian is unconcerned (i.e., $\beta=0$ and/or $Q_s^H(\cdot)=0$), the pedestrian keeps walking toward the target position $g_H$. Thus, the self-driving car comes to a full stop to keep the probability of collision lower than the threshold value. Whereas, when the robot alerts the danger to a concerned but unaware pedestrian, he/she runs backward to the right sidewalk. Thus the self-driving car continues toward the goal position $g_R$ without stopping. 
%Note that since the coefficient $\beta$ changes sometime during the interaction interval, the robot cannot identify the actual $\beta$ accurately. However, the robot recognizes the change in the opinion of the pedestrian. 

\begin{figure}[!t]
\centering
\includegraphics[width=7cm]{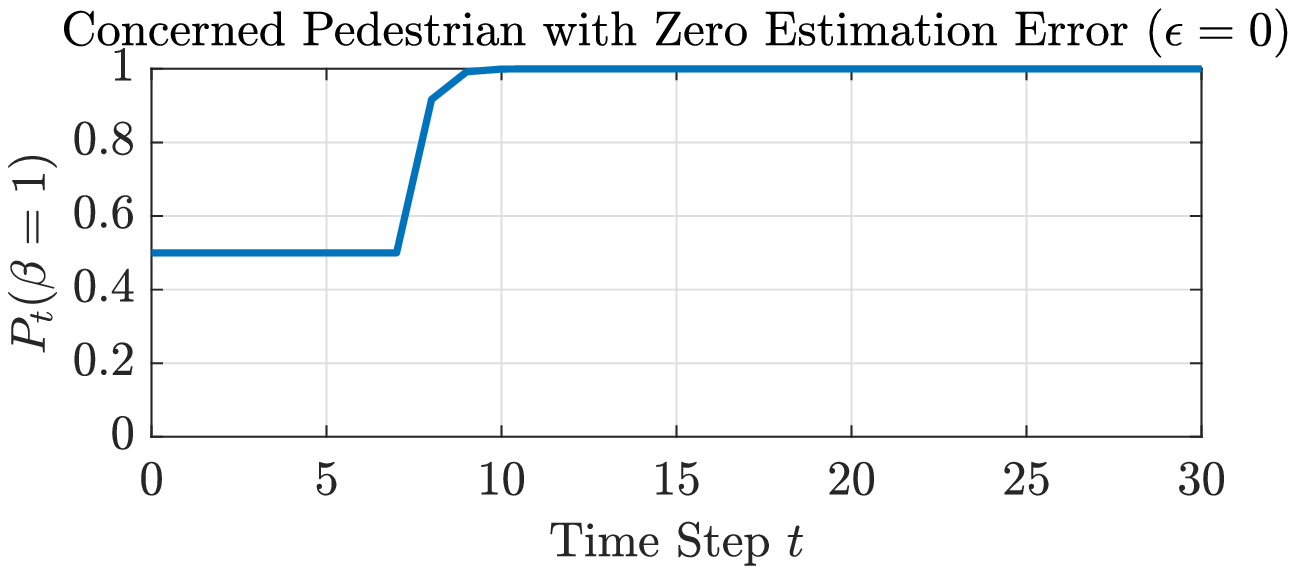}\\
\includegraphics[width=7cm]{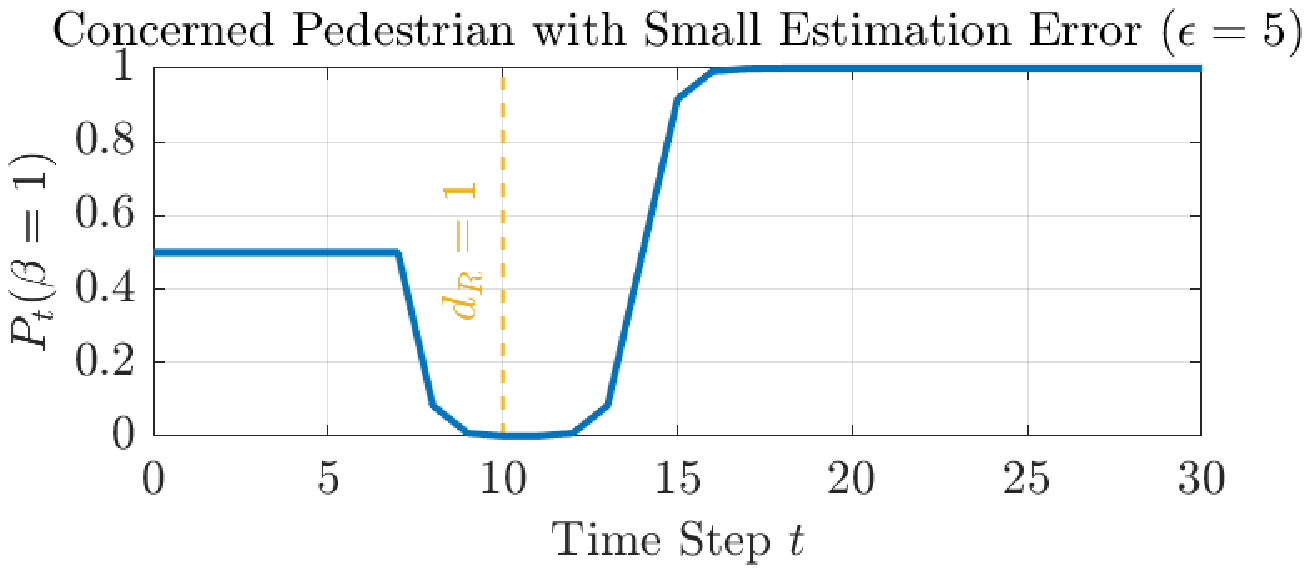}\\
\includegraphics[width=7cm]{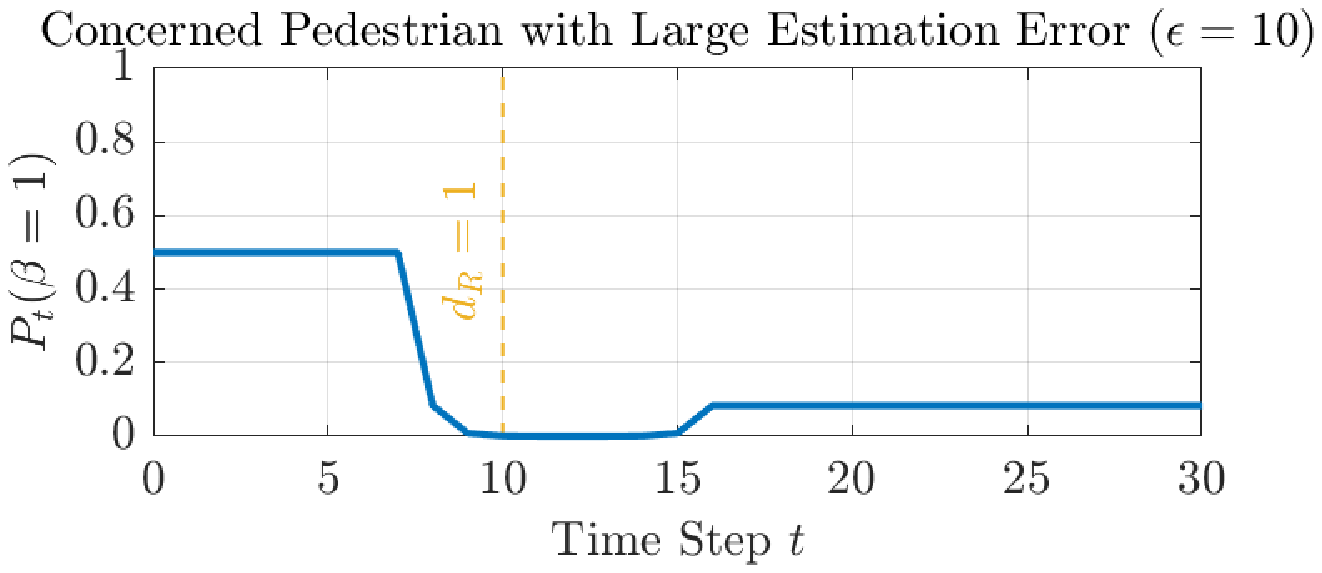}
\caption{Time profile of the robot's belief about the likelihood that the pedestrian is aware of the danger in the presence of a concerned pedestrian. Top figure: the estimation error is zero; the pedestrian realizes the danger in time. Middle figure: the pedestrian realizes the danger late, and runs toward the sidewalk on the right. Bottom figure: the pedestrian realizes the danger very late, and runs toward the sidewalk on the left.}
\label{fig:BeliefProfile1}
\end{figure}

\subsection{Impact of the Estimation Error $\epsilon[t]$}
As discussed in Section \ref{sec:HumanModel}, the pedestrian makes use of an estimation of the position of the self-driving car. However, as discussed in Section \ref{sec:HumanActionPrediction}, the self-driving car predicts the pedestrian's action based on its actual position. This means that in the presence of a large estimation error, even if the pedestrian is completely aware of the danger (i.e., $\beta=1$), the pedestrian may take an action which increases the risk. Thus, the self-driving car cannot learn accurately the danger awareness coefficient, and may have unnecessary stops. As discussed in Remark \ref{remark:dangernotifier}, the danger signaling system can impact the interaction by decreasing the estimation error. This impact has been studied in Fig. \ref{fig:BeliefProfile1}. As seen in this figure, even though the pedestrian is concerned, due to the estimation error the pedestrian takes a safe action late\footnote{We assume that the pedestrian compensates the estimation error as $\epsilon[t]=\epsilon_0\cdot e^{-\eta.d_R.(t-t_d)}$, where $\epsilon_0$ is the initial error, $\eta>0$ is a scalar, and $t_d$ is the time that $d_R$ switches from 0 to 1.}. Note that when the estimation error is small (i.e., $\epsilon[t]=5$), the pedestrian runs backward toward the sidewalk on the right, as the pedestrian realizes the danger when he/she is in the right half of the street. While, when the estimation error is large (i.e., $\epsilon[t]=10$), the pedestrian runs forward toward the sidewalk on the left, as the pedestrian realizes the danger when he/she is in the left half of the street.

\subsection{Impact of the Mixture Weight $\omega_H$}
\begin{figure}
\centering
\includegraphics[width=7cm]{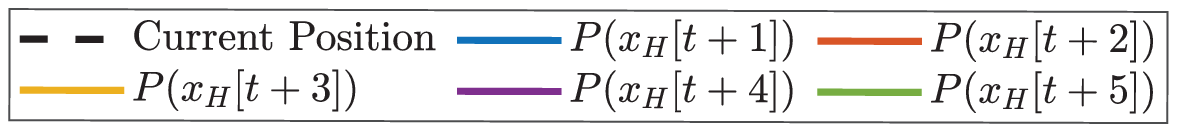}\\
\includegraphics[width=4.3cm]{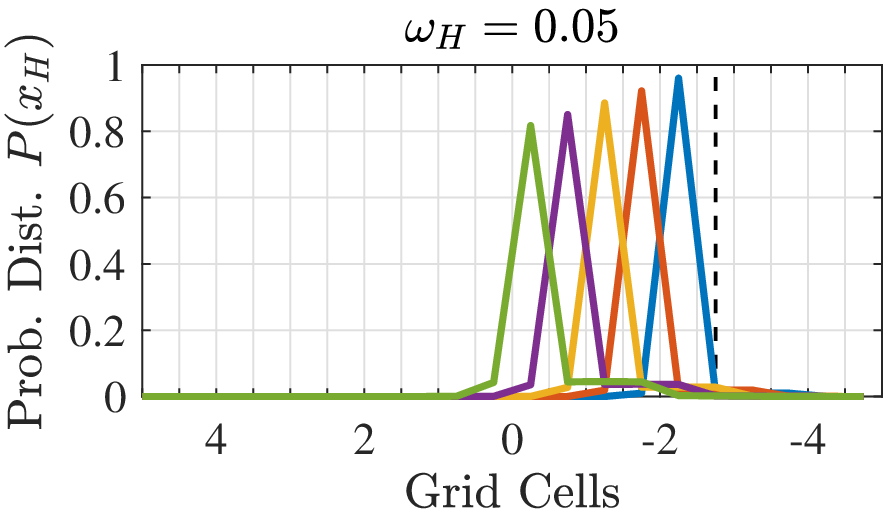}
\includegraphics[width=4.3cm]{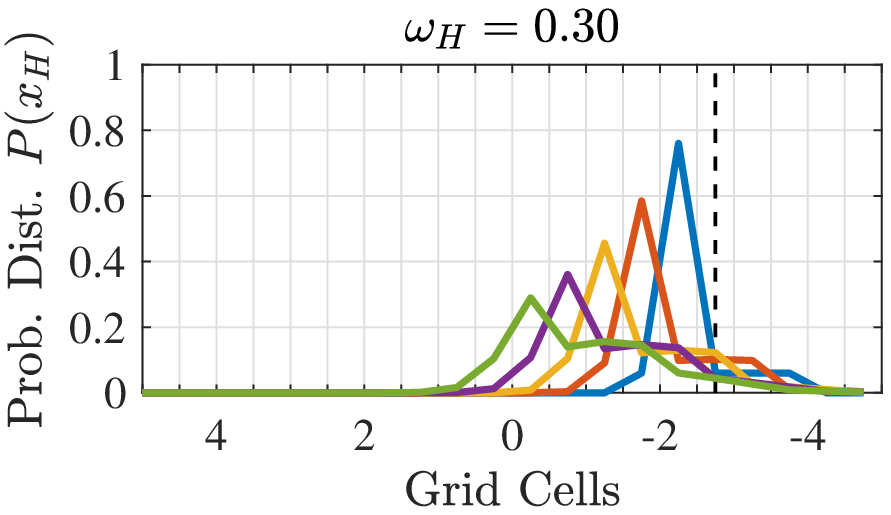}\\
\includegraphics[width=4.3cm]{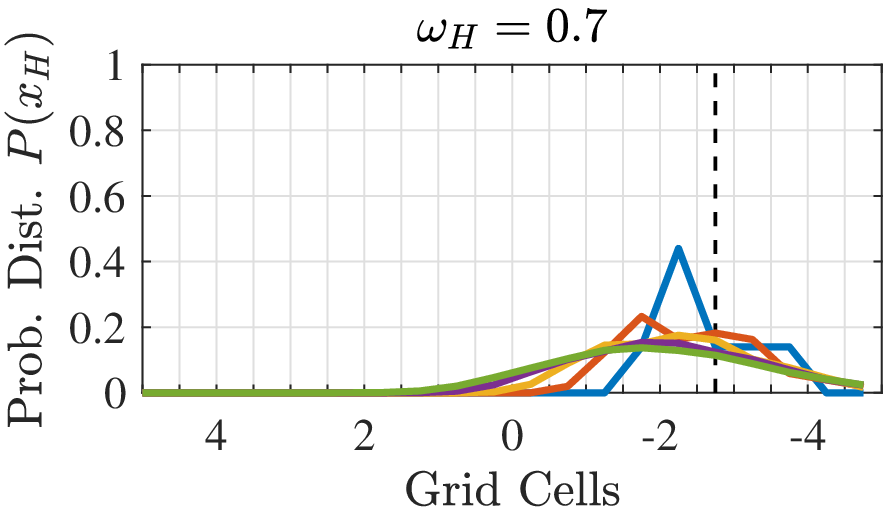}
\includegraphics[width=4.3cm]{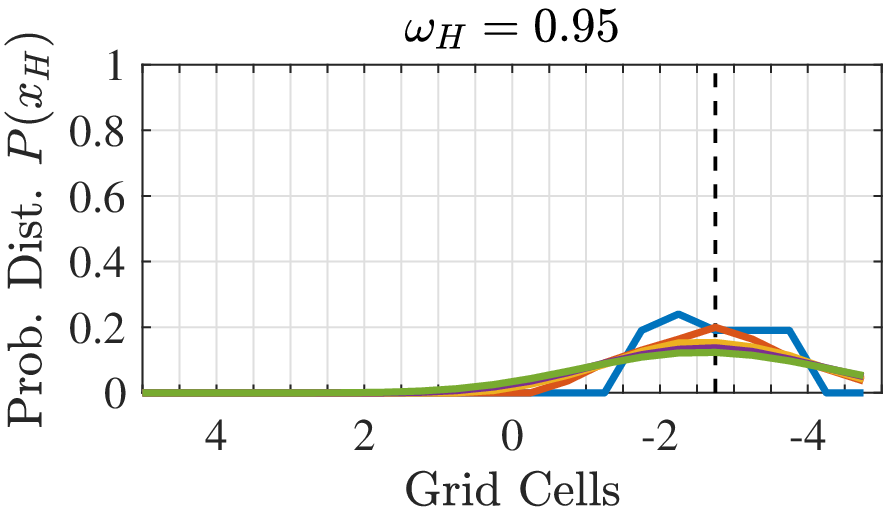}
\caption{The impact of the mixture weight $\omega_H$ in the probability distribution over the human's position in the street over the prediction horizon, computed at time $t$. Note that the pedestrian moves from right to left.}
\label{fig:EffectOmega}
\end{figure}

As discussed in Section \ref{sec:HumanActionPrediction}, the mixture weight $\omega_H$ defines the relationship between the mixture components. More precisely, $\omega_H=0$ means that the human is being driven only by the goal and safety objective functions, and $\omega_H=1$ means that the human chooses the actions randomly by completely ignoring the objectives functions. This mixture weight affects the prediction of the pedestrian's position in the street over the prediction horizon. 
In particular, for a large $\omega_H$, as the pedestrian appears random, the probability distribution over the pedestrian's position in the future will be wide. While, a small $\omega_H$ leads to a tight distribution. This impact is shown in Fig. \ref{fig:EffectOmega} for four different values of $\omega_H$. As seen in this figure, by increasing $\omega_H$, the probability distribution over the pedestrian's position in the street becomes wider, meaning that the prediction of the pedestrian's position becomes more uncertain.

%%%%%%%%%%%%%%%%%%%%%%
\section{Conclusion and Future Work}\label{sec:conclusion}
This paper introduced the notion of danger awareness in HRI, and accordingly the co-called danger awareness coefficient. This coefficient quantifies the human's intention to participate in and/or human's opinion of cooperative safety enforcement. The notion of danger awareness contributes to the state-of-the-art by revoking the presumption that humans do not intend to cooperate with robots to enforce safety, which usually leads to a conservative solution. In particular, this notion not only addresses Law\#1 and \#2 of the Isaac Asimov's ``Three Laws of Robotics" (i.e., safety-related laws), but also tackles Law\#3 (i.e., efficiency-related law). It is noteworthy that fully transferring safety responsibility to the robot while the human is eager to contribute goes against the efficiency of the robot.

This paper proposed an online Bayesian method to learn the value of the danger awareness coefficient. It was shown how this learning scheme helps the robot to build a predictive human model to predict the human's future actions. Since the learning scheme is based on real-time observations, this paper deployed an danger signaling system to actively perturb the environment and enrich the observations, and consequently improve the learning performance. 
It should be remarked that the danger signaling system also improves the efficiency of the robot by  acquainting an unaware human and driving the human to assist the robot in enforcing the safety.

Finally, a predictive planning scheme was proposed to obtain a plan for the robot. It was shown that by leveraging the danger awareness coefficient and the danger signaling system, the planning scheme obtains a probabilistically safe, but yet efficient, plan. The proposed scheme was verified through intensive simulation studies on an interaction between a self-driving car and a pedestrian, and the impact of different parameters were assessed.

Based upon our results in this paper, we believe that the notion of danger awareness has potential for playing a core role in enabling autonomous systems to have an efficient and safe interaction with humans. In particular, we envision exploiting this notion to prevent misuse of safety regulations by humans. By understanding the \textit{hidden intention} of humans, this paper revealed that it is possible to enforce safety without hindering the robots. In future work, we will investigate how the notion of danger awareness can be leveraged to improve \textit{fairness} in HRI despite conflicting objectives, i.e., ensuring safety of all agents without letting them deceive each other.

%%%%%%%%%%%%%%%%%%%%%%%%%%%%%
%\balance
\bibliographystyle{IEEEtran}
\bibliography{ref}

\begin{thebibliography}{10}
\providecommand{\url}[1]{#1}
\csname url@samestyle\endcsname
\providecommand{\newblock}{\relax}
\providecommand{\bibinfo}[2]{#2}
\providecommand{\BIBentrySTDinterwordspacing}{\spaceskip=0pt\relax}
\providecommand{\BIBentryALTinterwordstretchfactor}{4}
\providecommand{\BIBentryALTinterwordspacing}{\spaceskip=\fontdimen2\font plus
\BIBentryALTinterwordstretchfactor\fontdimen3\font minus
  \fontdimen4\font\relax}
\providecommand{\BIBforeignlanguage}[2]{{%
\expandafter\ifx\csname l@#1\endcsname\relax
\typeout{** WARNING: IEEEtran.bst: No hyphenation pattern has been}%
\typeout{** loaded for the language `#1'. Using the pattern for}%
\typeout{** the default language instead.}%
\else
\language=\csname l@#1\endcsname
\fi
#2}}
\providecommand{\BIBdecl}{\relax}
\BIBdecl

\bibitem{Brooks2019}
C.~Brooks and D.~Szafir, ``Balanced information gathering and goal-oriented
  actions in shared autonomy,'' in \emph{Proc. 14th ACM/IEEE Int. Conf.
  Human-Robot Interaction}, Daegu, Korea, Mar. 11-14, 2019, pp. 85--94.

\bibitem{Shamouty2020}
M.~El-Shamouty, X.~Wu, S.~Yang, M.~Albus, and M.~F. Huber, ``Towards safe
  human-robot collaboration using deep reinforcement learning,'' in \emph{Proc.
  Int. Conf. Robotics and Automation}, Paris, France, May 31-Aug. 31, 2020, pp.
  4899--4905.

\bibitem{Keil2020}
D.~Fridovich-Keil, A.~Bajcsy, J.~F. Fisac, S.~L. Herbert, S.~Wang, A.~D.
  Dragan, and C.~J. Tomlin, ``Confidence-aware motion prediction for real-time
  collision avoidance,'' \emph{The Int. J. Robotics Research}, vol.~39, no.
  2-3, pp. 250--265, Mar. 2020.

\bibitem{Peddi2020}
R.~Peddi, C.~D. Franco, S.~Gao, and N.~Bezzo, ``A data-driven framework for
  proactive intention-aware motion planning of a robot in a human
  environment,'' in \emph{Proc. IEEE/RSJ Int. Conf. Intelligent Robots and
  Systems}, Las Vegas, NV, USA, Oct. 25-29, 2020, pp. 5738--5744.

\bibitem{Huber2010}
M.~Huber, A.~Knoll, T.~Brandt, and S.~Glasauer, ``When to assist?-modelling
  human behaviour for hybrid assembly systems,'' in \emph{Proc. 41st Int. Symp.
  Robotics and 6th German Conf. Robotics}, Munich, Germany, Jun. 7-9, 2010.

\bibitem{Shi2004}
Y.~Shi, Y.~Huang, D.~Minnen, A.~Bobick, and I.~Essa, ``Propagation networks for
  recognition of partially ordered sequential action,'' in \emph{Proc. IEEE
  Computer Society Conf. Computer Vision and Pattern Recognition}, Washington,
  DC, USA, Jun. 27- Jul. 2, 2004, pp. 862--869.

\bibitem{Albanese2008}
M.~Albanese, R.~Chellappa, V.~Moscato, A.~Picariello, V.~S. Subrahmanian, and
  P.~Turaga, ``A constrained probabilistic petri net framework for human
  activity detection in video,'' \emph{IEEE Trans. Multimed.}, vol.~10, no.~6,
  pp. 982--996, Oct. 2008.

\bibitem{Kinugawa2017}
J.~Kinugawa, A.~Kanazawa, S.~Arai, and K.~Kosuge, ``Adaptive task scheduling
  for an assembly task coworker robot based on incremental learning of human
  motion patterns,'' \emph{IEEE Robot. Autom. Lett.}, vol.~2, no.~2, pp.
  856--863, Apr. 2017.

\bibitem{Vasquez2009}
D.~Vasquez, T.~Fraichard, and C.~Laugier, ``Incremental learning of statistical
  motion patterns with growing hidden markov models,'' \emph{IEEE Trans.
  Intell. Transp. Syst.}, vol.~10, no.~3, pp. 403--416, Sep. 2009.

\bibitem{Morris2011}
B.~T. Morris and M.~M. Trivedi, ``Trajectory learning for activity
  understanding: Unsupervised, multilevel, and long-term adaptive approach,''
  \emph{IEEE Trans. Pattern Anal. Mach. Intell.}, vol.~33, no.~11, pp.
  2287--2301, Nov. 2011.

\bibitem{Ding2011}
H.~Ding, G.~Reibig, K.~Wijaya, D.~Bortot, K.~Bengler, and O.~Stursberg, ``Human
  arm motion modeling and long-term prediction for safe and efficient
  human-robot-interaction,'' in \emph{Proc. IEEE Int. Conf. Robotics and
  Automation}, Shanghai, China, May 9-13, 2011, pp. 5875--5880.

\bibitem{Amor2014}
H.~B. Amor, G.~Neumann, S.~K. nd~Oliver~Kroemer, and J.~Peters, ``Interaction
  primitives for human-robot cooperation tasks,'' in \emph{Proc. IEEE Int.
  Conf. Robotics and Automation}, Hong Kong, China, May 31-Jun. 7, 2014, pp.
  2831--2837.

\bibitem{Koppula2013}
H.~S. Koppula and A.~Saxena, ``Anticipating human activities for reactive
  robotic response,'' in \emph{Proc. IEEE/RSJ Int. Conf. Intelligent Robots and
  Systems}, Tokyo, Japan, Nov. 3-7, 2013, p. 2071.

\bibitem{Li2012}
K.~Li, J.~Hu, and Y.~Fu, ``Modeling complex temporal composition of actionlets
  for activity prediction,'' in \emph{Proc. European Conference on Computer
  Vision}, Florence, Italy, Oct. 7-13, 2012, pp. 286--299.

\bibitem{Baker2007}
C.~L. Baker, J.~B. Tenenbaum, and R.~R. Saxe, ``Goal inference as inverse
  planning,'' \emph{Proceedings of the Annual Meeting of the Cognitive Science
  Society}, vol.~29, pp. 779--784, 2007.

\bibitem{Neumann2007}
J.~V. Neumann and O.~Morgenstern, \emph{Theory of games and economic
  behavior}.\hskip 1em plus 0.5em minus 0.4em\relax Princeton University Press,
  2007.

\bibitem{Luce2012}
R.~D. Luce, \emph{Individual Choice Behavior: A Theoretical Analysis}.\hskip
  1em plus 0.5em minus 0.4em\relax Courier Corporation, 2012.

\bibitem{Ziebart2009}
B.~D. Ziebart, N.~Ratliff, G.~Gallagher, C.~Mertz, K.~Peterson, J.~A. Bagnell,
  M.~Hebert, A.~K. Dey, and S.~Srinivasa, ``Planning-based prediction for
  pedestrians,'' in \emph{Proc. IEEE/RSJ Int. Conf. Intelligent Robots and
  Systems}, St. Louis, MO, USA, Oct. 10-15, 2009, pp. 3931--3936.

\bibitem{Fisac2018}
J.~F. Fisac, A.~Bajcsy, S.~Herbert, D.~Fridovich-Keil, S.~Wang, C.~J. Tomlin,
  and A.~D. Dragan, ``Probabilistically safe robot planning with
  confidence-based human predictions,'' in \emph{Proc. Robotics: Science and
  Systems}, Pittsburgh, PA, USA, Jun. 26-30, 2018.

\bibitem{Bajcsy2019}
A.~Bajcsy, S.~L. Herbert, D.~Fridovich-Keil, J.~F. Fisac, S.~Deglurkar, A.~D.
  Dragan, and C.~J. Tomlin, ``A scalable framework for real-time multi-robot,
  multi-human collision avoidance,'' in \emph{Proc. Int. Conf. Robotics and
  Automation}, Montreal, QC, Canada, May 20-24, 2019, pp. 936--943.

\bibitem{Hawkins2018}
K.~P. Hawkins and P.~Tsiotras, ``Anticipating human collision avoidance
  behavior for safe robot reaction,'' in \emph{Proc. IEEE Conf. Decision and
  Control}, Miami Beach, FL, USA, Dec. 17-19, 2018, pp. 6301--6306.

\bibitem{Wilcox2012}
R.~Wilcox, S.~Nikolaidis, and J.~Shah, ``Optimization of temporal dynamics for
  adaptive human-robot interaction in assembly manufacturing,'' in \emph{Proc.
  Robotics: Science and Systems}, Sydney, NSW, Australia, Jul. 9-13, 2012, pp.
  441--448.

\bibitem{Ding2014}
H.~Ding, M.~Schipper, and B.~Matthias, ``Optimized task distribution for
  industrial assembly in mixed human-robot environments-- {C}ase study on {IO}
  module assembly,'' in \emph{Proc. IEEE Int. Conf. Automation Science and
  Engineering}, Taipei, Taiwan, Aug. 18-22, 2014, pp. 19--24.

\bibitem{Hawkins2013}
K.~P. Hawkins, N.~Vo, S.~Bansal, and A.~F. Bobick, ``Probabilistic human action
  prediction and wait-sensitive planning for responsive human-robot
  collaboration,'' in \emph{Proc. 13th IEEE-RAS Int. Conf. Humanoid Robots},
  Atlanta, GA, USA, Oct. 15-17, 2013, pp. 499--506.

\bibitem{Hawkins2014}
K.~P. Hawkins, S.~Bansal, N.~N. Vo, and A.~F. Bobick, ``Anticipating human
  actions for collaboration in the presence of task and sensor uncertainty,''
  in \emph{Proc. IEEE Int. Conf. Robotics and Automation}, Hong Kong, China,
  May 31- Jun. 7, 2014, pp. 2215--2222.

\bibitem{Tanaka2012}
Y.~Tanaka, J.~Kinugawa, and K.~Kosuge, ``Motion planning with worker's
  trajectory prediction for assembly task partner robot,'' in \emph{Proc.
  IEEE/RSJ Int. Conf. Intelligent Robots and Systems}, Vilamoura, Portugal,
  Oct. 7-12, 2012, pp. 1525--1532.

\bibitem{Kanazawa2019}
A.~Kanazawa, J.~Kinugawa, and K.~Kosuge, ``Adaptive motion planning for a
  collaborative robot based on prediction uncertainty to enhance human safety
  and work efficiency,'' \emph{IEEE Trans. Robot.}, vol.~35, no.~4, pp.
  817--832, Aug. 2019.

\bibitem{Baizid2015}
K.~Baizid, A.~Yousnadj, A.~Meddahi, R.~Chellali, and J.~Iqbal, ``Time
  scheduling and optimization of industrial robotized tasks based on genetic
  algorithms,'' \emph{Robot. Comput.-Integr. Manuf.}, vol.~34, pp. 140--150,
  Aug. 2015.

\bibitem{Belkhouche2009}
F.~Belkhouche, ``Reactive path planning in a dynamic environment,'' \emph{IEEE
  Trans. Robotics}, vol.~25, no.~4, pp. 902--911, Aug. 2009.

\bibitem{Aoude2013}
G.~S. Aoude, B.~D. Luders, J.~M. Joseph, N.~Roy, and J.~P. How,
  ``Probabilistically safe motion planning to avoid dynamic obstacles with
  uncertain motion patterns,'' \emph{Autonomous Robots}, vol.~35, pp. 51--76,
  2013.

\bibitem{HosseinzadehICRA}
M.~Hosseinzadeh, B.~Sinopoli, and A.~F. Bobick, ``Toward a safe and efficient
  human-robot interaction: When you have a careless worker!'' in \emph{Proc.
  Int. Conf. Robotics and Automation}, Xi'an, China, May 30-Jun. 5, 2021.

\bibitem{Luce2005}
R.~D. Luce, \emph{Individual Choice Behavior: A Theoretical Analysis}.\hskip
  1em plus 0.5em minus 0.4em\relax Dover Publications, Inc., 2005.

\bibitem{Baraka2016}
K.~Baraka, S.~Rosenthal, and M.~Veloso, ``Enhancing human understanding of a
  mobile robot's state and actions using expressive lights,'' in \emph{Proc.
  25th IEEE Int. Symp. Robot and Human Interactive Communication}, New York,
  NY, USA, Aug. 26-31, 2016, pp. 652--657.

\bibitem{Wogalter2018}
M.~S. Wogalter, ``Communication-human information processing (c-hip) model in
  forensic warning analysis,'' in \emph{Proc. 20th Congress Int. Ergonomics
  Association}, Florence, Italy, Aug. 26-30, 2018, pp. 761--769.

\bibitem{Zacharias2001}
J.~Zacharias, ``Pedestrian behavior and perception in urban walking
  environments,'' \emph{J. Planning Literature}, vol.~16, no.~1, pp. 3--18,
  Aug. 2001.

\bibitem{Campbell2003}
B.~J. Campbell, C.~V. Zegeer, H.~H. Huang, and M.~J. Cyneck, ``{A Review of
  Pedestrian Safety Research in the United States and Abroad},'' University of
  North Carolina, Tech. Rep., 11 2003.

\bibitem{Mehta2008}
V.~Mehta, ``Walkable streets: pedestrian behavior, perceptions and attitudes,''
  \emph{J. Urbanism: Int. Research on Placemaking and Urban Sustainability},
  vol.~1, no.~3, pp. 217--245, 2008.

\bibitem{McAslan2017}
D.~McAslan, ``Walking and transit use behavior in walkable urban
  neighborhoods,'' \emph{Michigan J. Sustainability}, vol.~5, no.~1, pp.
  51--71, 2017.

\bibitem{Chapman2019}
M.~P. Chapman, J.~Lacotte, A.~Tamar, D.~Lee, K.~M. Smith, V.~Cheng, J.~F.
  Fisac, S.~Jha, M.~Pavone, and C.~J. Tomlin, ``A risk-sensitive finite-time
  reachability approach for safety of stochastic dynamic systems,'' in
  \emph{Proc. American Control Conf.}, Philadelphia, PA, USA, Jul. 10-12, 2019.

\bibitem{yalmip}
J.~Lofberg, ``{YALMIP}: a toolbox for modeling and optimization in {MATLAB},''
  in \emph{Proc. Int. Conf. Robotics and Automation}, New Orleans, LA, USA,
  Sep. 2-4, 2004, pp. 284--289.

\end{thebibliography}

\end{document}